\documentclass{article}

\usepackage{PRIMEarxiv}

\usepackage{fontspec}
\usepackage{xeCJK}

\IfFontExistsTF{IPAexMincho}
  {\setCJKmainfont{IPAexMincho}}
  {\setCJKmainfont{HaranoAjiMincho}}

\usepackage{xcolor}
\usepackage[
    colorlinks=true,
    linkcolor=blue,
    citecolor=blue,
    urlcolor=blue
]{hyperref}
\usepackage{url}
\usepackage{booktabs}
\usepackage{amsfonts}
\usepackage{nicefrac}
\usepackage{microtype}
\usepackage{fancyhdr}
\usepackage{graphicx}
\usepackage{natbib}
\setcitestyle{numbers,square}
\usepackage{doi}
\usepackage{tabularx}
\usepackage{array}
\usepackage{float}
\usepackage{comment}
\usepackage[page]{appendix}

\graphicspath{{media/}{./}}

\title{Authorship attribution and aesthetic evaluation of AI poetry: a case study with Haiku}

\author{
  Livia Oddi$^{1,2}$ \\
  DIET, Sapienza University of Rome, Rome, Italy \\
  Shibaura Institute of Technology, Tokyo, Japan \\
  \texttt{oddi.1846085@studenti.uniroma1.it}
  \And
  Simone Scardapane$^{1}$ \\
  DIET, Sapienza University of Rome, Rome, Italy
  \And
  Toru Sugimoto$^{2}$ \\
  Shibaura Institute of Technology, Tokyo, Japan
  \And
  Donatella Genovese$^{3}$ \\
  DIAG, Sapienza University of Rome, Rome, Italy
}

\hypersetup{
    pdftitle={Authorship Attribution and Aesthetic Evaluation of AI Poetry: A Case Study with Haiku},
    pdfauthor={Livia Oddi, Simone Scardapane, Toru Sugimoto, Donatella Genovese},
    pdfkeywords={large language models, computational creativity, haiku, authorship attribution, human evaluation, Japanese NLP}
}

\begin{document}
\maketitle

\begin{abstract}
This paper investigates the generation and human evaluation of Japanese haiku by contemporary Large Language Models (LLMs), focusing on authorship perception and aesthetic judgment within a constrained poetic form. Using a few-shot prompting strategy, Japanese haiku were generated across a heterogeneous set of large language models, including open- and closed-source systems, medium-scale and large-scale architectures, models with native or adapted Japanese support, and multilingual proprietary models. These AI-generated haiku were combined with human-written ones and presented in a questionnaire distributed to students at Japanese universities in Tokyo.
The survey assessed whether respondents could distinguish between AI-generated and human-written haiku and which cues informed their judgments. Recognition accuracy varied across models. GPT-5, Gemini 2.5, and StableLM-7B performed at approximately chance level ($\approx 0.50$), whereas LLM-JP, Gemma-2B, and LLaMA-2 showed moderate detectability ($\approx 0.59$--$0.67$). However, recognition was strongly item-dependent. Ratings of fluency, coherence, poeticness, and related aesthetic dimensions predicted perceived humanness but not correct classification, indicating an attribution bias linked to aesthetic evaluation and revealing a dissociation between aesthetic evaluation and true authorship detection.
The extended analysis additionally examines generation-constraint adherence, participant-level characteristics, and exploratory LLM-based evaluations of haiku authorship. Overall, the findings suggest that as LLMs improve, surface-level creative plausibility may reduce reliable human discrimination within constrained poetic settings.
\end{abstract}


\section{Introduction}

Research on computational creativity has a long history, but interest in evaluating AI creativity has intensified significantly following the emergence of Large Language Models (LLMs), particularly in generative systems that operate in natural language (e.g., \cite{mizrahi2025cooking}). LLMs architectures are based on the Transformer architecture introduced by \cite{vaswani2017attention}, which uses self-attention mechanisms to efficiently model contextual dependencies at scale. Beyond pretraining scale alone, recent improvements in LLM performance have increasingly come from instruction tuning, alignment procedures, and reasoning-oriented post-training methods \cite{openai2025gpt5, electronics14183580}. These developments are particularly relevant for constrained creative generation tasks such as haiku, where the model must not only produce fluent Japanese text, but also follow explicit formal requirements, including seasonal reference, three-line structure, and 5–7–5 mora constraints. In this context, instruction-following ability becomes central, because successful generation depends on the model’s capacity to satisfy stylistic, linguistic, and structural constraints simultaneously.%Building on this architecture, models such as \textbf{BERT} \cite{devlin2019bert} introduced bidirectional contextual representations trained with masked language modeling objectives, enabling strong performance in language understanding tasks. Subsequently, \textbf{GPT} \cite{radford2018improving} and other decoder-based models adopted an autoregressive training objective, focusing on next-token prediction to generate coherent and contextually consistent text. Larger-scale systems such as \textbf{PaLM} \cite{chowdhery2022palm} and \textbf{LLaMA} \cite{touvron2023llama} further extended this paradigm by scaling model parameters and training data, demonstrating strong generalization capabilities across a wide range of natural language tasks \cite{electronics14183580}.

As generative AI systems gradually become part of everyday life, the question is no longer whether they can produce human-like outputs, but rather whether they can generate creations that humans judge creatively meaningful. Empirical studies comparing human-made and AI-generated poetry have begun to explore this issue \cite{hitsuwari2023haiku}. These technologies now extend across nearly every domain, including the arts and other creative disciplines, which are traditionally considered deeply human \cite{electronics14183580}. Poetry is a particularly interesting case, as it combines formal constraints, cultural conventions, and subtle aesthetic judgment, making it an informative domain for examining human evaluation and authorship perception in AI-generated poetry. \cite{hitsuwari2023haiku}.

A particularly interesting form of poetry is haiku, a type of traditional Japanese poetic form structured around seasonal reference (kigo), brevity, and a cutting moment (kire). Formally, haiku can be regular, with a 5-7-5 mora\footnote{Timing units that do not always coincide with syllables.} structure, or irregular. Focusing on regular haiku as a case study allows us to have a controlled generation space, with a constrained output comparable across models while still leaving room for creativity. This combination of constraints and creative freedom makes haiku a suitable testbed for examining human evaluation and authorship perception in AI-generated poetry.

Although prior studies have examined the aesthetic evaluation of AI-generated poetry and compared human- and machine-authored texts \cite{hitsuwari2023haiku}, and other research has focused on improving generative creativity in large language models \cite{mizrahi2025cooking}, relatively little attention has been devoted to a comparative analysis that simultaneously evaluates the linguistic quality of generated haiku across multiple contemporary models and the relationship between perceived creativity and correct authorship attribution. In particular, while previous work has explored either aesthetic judgments \cite{CHIARELLA2022107406,BARA2025106063} or generative enhancement techniques \cite{mizrahi2025cooking,peeperkorn2024temperature}, a systematic cross-model investigation within a formally constrained setting remains limited.This raises two related questions: whether the cues driving perceived humanness and creativity in generated poetry also enable correct authorship attribution, and whether these evaluative patterns remain stable across different generative systems.%It is therefore still unclear whether the cues that drive perceived humanness and creativity in generated text are also diagnostic for correct authorship attribution, and how robust these judgments are across different generative systems.%

To address this gap, the present study investigates how modern LLMs generate Japanese haiku within a constrained prompting framework, and how humans perceive, evaluate, and attribute authorship to these outputs. The haiku generation process employed a few-shot prompting technique using examples from the Modern Haiku Dataset \href{https://huggingface.co/datasets/p1atdev/modern_haiku}{p1atdev}, which contains 37,158 haiku in JSONL format written by 4,625 Japanese poets. Three regular haiku were generated for each season (spring, summer, autumn, and winter) using six different AI models (GPT-5, Gemini 2.5, Gemma-2B, StableLM-7B, LLaMA-2 and LLM-JP) across open- and closed-source systems and different model sizes. These were then mixed with three human-written haiku per season, sourced from online haiku contests, and presented to 144 Japanese university students in a questionnaire. The students were asked to identify whether each haiku was written by AI or a human, and to evaluate various aspects of it (e.g. fluency, coherence, haiku-like wording) on a scale from 1 to 5. Importantly, the authorship-attribution task was not treated as a Turing test in which indistinguishability is taken as evidence of creativity. Rather, it was used to examine whether perceived authorship, aesthetic evaluation, and correct source recognition converge or diverge. The findings indicate that, for some of the strongest models, participants' authorship judgments are close to chance, suggesting that discrimination is weak and highly dependent on individual items rather than reflecting a stable model-level signal. At the same time, impression ratings are more closely tied to what participants believe the source is than to the true source. Overall, the results suggest a distinction between aesthetic impression and correct authorship identification.

More specifically, this study makes three key contributions. Firstly, it evaluates a diverse range of contemporary Large Language Models, including open- and closed-source systems, models with different parameter scales and varying degrees of Japanese-language specialization. Second, evaluation was conducted with Japanese university participants in a culturally grounded poetic setting involving seasonal fixed-form haiku rather than in English-language or externally sourced settings. Third, the study simultaneously examines the relationships between aesthetic evaluation, perceived authorship and correct recognition, enabling us to investigate whether the cues that drive perceived humanness also support accurate authorship attribution.

The findings should therefore be interpreted within the specific context of constrained Japanese haiku generation under a few-shot prompting framework, rather than as a general evaluation of AI creativity across unconstrained creative domains.

\section{State of the Art}

\subsection{Computational Poetry and Haiku Generation}

Poetry generation predates the era of neural language models. Earlier computational creativity systems explored rule-based and template-driven poetic composition, including haiku-oriented systems such as Gabriel's \textit{Inkwell} \cite{gabrielinkwell}, often cited as an early example of computational haiku generation. These systems typically emphasized structural control, formal constraints, and handcrafted design principles rather than large-scale statistical language modelling. Before the era of LLMs, neural approaches to poetry generation included recurrent architectures with iterative refinement. One example is Yan's iPoet system, which models generation as a process that can be refined over multiple steps \cite{yan2016ipoet}.
Other studies have explored the use of adversarial objectives for creative text generation, including the generation of haiku using Sequence Generative Adversarial Networks (SeqGAN) style methods \cite{hirota2018seqgan}.
More recent model-focused work investigates haiku generation with deep neural networks and autoregressive language models, including approaches tailored to seasonal, fixed-form Japanese haiku \cite{wu2017haiku, hirata2023haiku}.

\subsection{Evaluating Haiku: Automatic Scoring, Human Ratings, and LLM-as-Judge}

Alongside generation, a parallel line of research has focused on evaluation. Kikuchi et al. propose a neural approach to automatically estimating haiku quality using proxy signals (e.g. popularity/likes), demonstrating an early attempt to operationalise aesthetic judgment for Japanese haiku \cite{kikuchi2016haiku}.
Human evaluation remains central to many studies, including those comparing aesthetic responses to AI-generated poetry with those to poetry involving human intervention.
For example, Hitsuwari et al. (2023) conducted a controlled experiment in which participants rated human-made, AI-generated, and human-in-the-loop collaborative haiku along multiple aesthetic dimensions and also judged authorship. They reported that collaborative haiku received the highest beauty scores and that participants could not accurately discriminate between human and AI authorship \cite{hitsuwari2023haiku}.
More recently, Tomizawa proposed a method for evaluating haiku using a large language model (LLM) and prompt engineering. This method treats the LLM as an evaluator and compares its selections with human or community assessments (e.g. contest-related signals) \cite{tomizawa2025haiku}.
Together, these approaches demonstrate that haiku evaluation can be conducted using automatic proxies, direct human ratings and LLM-based evaluators. However, these approaches tend to prioritise either quality estimation or preference over the relationship between perceived humanness, aesthetic impressions and the ability to correctly attribute authorship in a controlled setting.
Although computational creativity theory offers conceptual distinctions, such as routes to novelty and the importance of valuation, and prior empirical studies \cite{CHIARELLA2022107406,BARA2025106063} have emphasised the significance of framing and authorship cues in aesthetic judgement, research on haiku remains divided between generation-focused and evaluation-focused studies. In particular, little work has been done to compare multiple contemporary generative models within the context of a formally constrained, culturally meaningful poetic form such as seasonal fixed-form Japanese haiku, or to disentangle how human evaluative cues relate to perceived humanness versus correct authorship attribution. Controlled experimental designs in which AI- and human-written haiku are mixed, rated along multiple linguistic and aesthetic dimensions, and judged for authorship enable a joint analysis of aesthetic evaluation and recognition performance, addressing this gap.

\section{Models and Generation Framework}

As this study examines LLM-based haiku generation and human evaluation within a constrained poetic setting, it was first necessary to construct a controlled generation framework.
Haiku were employed as a constrained poetic testbed, enabling the examination of creativity within a form that combines rigid structural rules with scope for aesthetic variation. The generation process was designed not only to produce plausible poems, but also to ensure comparability across models while maintaining key formal haiku properties, such as the 5-7-5 mora structure and the presence of seasonal references (kigo).
To support this objective, we developed a pipeline that combines a curated haiku corpus with few-shot prompting strategies applied to several contemporary language models.

\subsection{Dataset and Preprocessing}

We used the haiku corpus p1atdev, comprising 37,158 Japanese haiku written by 4,625 poets in JSONL format. Each entry includes the poem text together with metadata such as author, source, season, and an embedded \emph{kigo} dictionary. The \emph{kigo} metadata includes the seasonal word itself, its pronunciation (\texttt{kana}), an optional old orthographic form (\texttt{old\_kana}), the associated season, and possible alternative names (\texttt{subtitle}).
During preprocessing, no haiku records were removed at the row level: the full set of 37,158 poems was retained (100\%), and 0\% of poem entries were excluded. Instead, preprocessing consisted of cleaning sparse metadata fields. In particular, the columns \texttt{foreword}, \texttt{source}, \texttt{comment}, \texttt{reviewer}, and \texttt{note} were dropped because they contained missing values for nearly the entire dataset and did not provide information necessary for generation or evaluation. In this paper, we use the term \emph{non-poetic metadata} to refer to such auxiliary editorial or archival fields, rather than to the haiku text itself or its seasonal annotation.
Each haiku was then labeled as regular or irregular using a mora-counting function. Poems matching the canonical 5-7-5 mora structure were labeled as regular, whereas poems deviating from this pattern were labeled as irregular.
The built-in \emph{kigo} dictionary was not created externally, but extracted directly from the dataset's native \texttt{kigo} field. During preprocessing, the \texttt{kigo} field was separated from the main poem dataframe into a dedicated \texttt{kigo\_df}, while the main corpus was stored as \texttt{train\_df} without the \texttt{kigo} column. This separation was introduced to allow explicit seasonal control during prompting and to facilitate the identification of seasonal words during tokenization. Within the \emph{kigo} metadata, the \texttt{old\_kana} field was removed because it contained a high number of null values, whereas \texttt{subtitle} was retained because it provided potentially useful additional seasonal variants.
Archaic kana characters such as ゑ and ゐ were retained rather than normalized, as they reflect historical orthography and preserve stylistic authenticity.

\subsection{Models}

A set of Japanese-capable large language models spanning open- and closed-source systems and different parameter scales was evaluated. The selected systems support Japanese through native, adapted, or multilingual training and were used with their original pretrained weights and built-in tokenizers. No external tokenization pipeline was employed during generation.
The models were selected to compare medium-scale open-source instruction-tuned systems, larger closed-source proprietary systems, and varying degrees of Japanese-language specialization and reasoning capability. Table~\ref{tab:models_profile} summarizes their main technical and institutional characteristics.

\begin{table}[H]
\centering
\caption{Technical and institutional profile of the evaluated models.}
\label{tab:models_profile}

\footnotesize
\setlength{\tabcolsep}{4pt}

\begin{tabular}{lllll}
\toprule
Model & Scale & Source & Japanese Support & Developer \\
\midrule

\href{https://huggingface.co/stabilityai/japanese-stablelm-instruct-gamma-7b}{StableLM-7B}
& 7B & Open & Native & Stability AI \\

\href{https://huggingface.co/elyza/ELYZA-japanese-Llama-2-7b}{LLaMA-2}
& 7B & Open & Adapted & ELYZA \\

\href{https://huggingface.co/google/gemma-2-2b-jpn-it}{Gemma-2B}
& 2B & Open & Adapted & Google \\

\href{https://huggingface.co/llm-jp/llm-jp-3.1-1.8b-instruct4}{LLM-JP}
& 1.8B & Open & Native & LLM-jp \\

\href{https://cdn.openai.com/gpt-5-system-card.pdf}{GPT-5}
& Proprietary & Closed & Multilingual & OpenAI \\

\href{https://deepmind.google/technologies/gemini/}{Gemini 2.5}
& Proprietary & Closed & Multilingual & Google \\

\bottomrule
\end{tabular}

\vspace{2mm}

\footnotesize
\textit{Note:} Model names are abbreviated for readability. The corresponding model releases are:
StableLM-7B (Japanese StableLM Instruct Gamma 7B),
LLaMA-2 (ELYZA-japanese-Llama-2-7B),
Gemma-2B (Gemma-2-2B-JPN-IT),
LLM-JP (LLM-JP-3.1-1.8B-Instruct4).

\end{table}
\subsection{Prompt Design}

Haiku generation was performed through few-shot prompting rather than parameter fine-tuning. This choice was motivated by empirical observations from preliminary experiments. Fine-tuning attempts, including hyperparameter optimization with Optuna and LoRA-based adaptation, consistently resulted in overfitting despite strong convergence in training loss. Even after restricting the dataset to structurally regular 5-7-5 haiku, the models tended to memorize training patterns rather than generalize the underlying compositional constraints. Additional experiments with smaller models suggested that this behavior was not solely attributable to model scale, but was also influenced by the limited size and high structural regularity of the dataset relative to the parameter space. Moreover, few-shot prompting was also the only viable option for the closed-source models, since proprietary systems such as GPT-5 and Gemini 2.5 cannot be fine-tuned or directly modified within the same experimental framework. By contrast, few-shot prompting provided a more stable and controllable generation framework. It allowed explicit conditioning on structure and seasonal information without modifying model parameters, reducing the risk of memorization while preserving generalization. Furthermore, prompting made it possible to directly control stylistic and formal constraints (e.g., mora structure and kigo inclusion) and to iterate rapidly on generation strategies, which was essential for comparing multiple models under consistent conditions.

Prompts were written in Japanese and required a three-line output, 5–7–5 mora structure, inclusion of at least one seasonal word (kigo), and the absence of Latin characters, numerical digits, ASCII punctuation, and other non-Japanese annotations. Few-shot conditioning was implemented by embedding example haiku in the prompt along with explicit metadata: 季語 (kigo), 季節 (season), 構造 (Regular / 5-7-5). This structured format encouraged explicit structural adherence.

An empirical effect emerged regarding prompt identity framing. When the identity line used a language-framed formulation (e.g., 「あなたは日本語の俳人です」) rather than a nationality-framed version (e.g., 「あなたは日本人の俳人です」), generation quality improved under identical decoding settings. For StableLM-7B, average perplexity decreased from approximately 52 to 34, accompanied by more coherent imagery and stronger seasonal alignment. A similar pattern was observed for Gemma-2B, where perplexity dropped from roughly 6975 to 530, alongside qualitative improvements in tone and haiku-like structure. Although the nationality framing is more idiomatic in Japanese, the language-framed formulation likely aligns more closely with common instruction patterns in model training data (e.g., ``do X in Japanese''), thereby acting as a stronger stylistic anchor. Given the consistent quantitative and qualitative gains, the language-framed formulation was adopted throughout the study.

\subsection{Decoding Strategy}

The decoding strategy plays an important role in creative text generation, as it influences the balance between structural coherence, lexical diversity, and originality. In the context of haiku generation, this trade-off is particularly relevant because the poems must satisfy strict formal constraints while still producing varied and aesthetically plausible outputs. In this study, generation was performed using stochastic sampling rather than beam search. Although beam search improved local likelihood and structural adherence, it frequently reused tokens and n-grams from few-shot examples, reducing lexical novelty and increasing prompt copying. This behavior is consistent with prior findings showing that beam search tends to produce generic and repetitive outputs, often favoring high-probability continuations at the expense of diversity \cite{holtzman2020curious, vijayakumar2016diverse}. Since haiku generation requires originality in addition to formal constraint satisfaction, sampling was preferred. A moderate temperature value was selected to balance novelty and coherence. Prior work on temperature and creativity in LLMs suggests that temperature is weakly correlated with novelty and moderately correlated with incoherence, while showing no relationship with cohesion or typicality \citep{peeperkorn2024temperature}. Therefore, temperature was not treated as a direct creativity parameter, but as a controlled stochastic setting intended to increase variation without excessively compromising coherence.

The final generation settings adopted in the experiments are summarized in Table \ref{tab:decoding}.

\begin{table}[H]
\centering
\caption{Final generation configuration (line-by-line stochastic decoding)}
\label{tab:decoding}
\footnotesize
\setlength{\tabcolsep}{4pt}
\begin{tabular}{lll}
\toprule
\textbf{Parameter} & \textbf{Value} & \textbf{Role} \\
\midrule
few\_shot\_k & 6 & Number of few-shot examples in prompt \\
num\_targets & 24 & Number of target haiku generated \\
max\_new\_tokens & 22 & Maximum tokens per generated line \\
do\_sample & True & Enables stochastic decoding \\
temperature & 0.7 & Controls randomness of sampling \\
top\_p & 0.90 & Nucleus sampling threshold \\
top\_k & 50 & Limits candidate token pool \\
repetition\_penalty & 1.15 & Penalizes repeated tokens \\
\bottomrule
\end{tabular}
\end{table}

Generation was performed using a line-by-line decoding strategy with stochastic sampling. Each haiku was generated sequentially (5-7-5), conditioning each line on the previously generated ones. This approach allowed strict enforcement of mora structure and controlled inclusion of the target seasonal word (kigo). Sampling was preferred over beam search to reduce prompt copying and increase lexical diversity. Although iterative conditioning occasionally reduced inter-line semantic coherence, it provided more reliable structural control, which was prioritized for evaluation. To construct the final evaluation set, more candidate haiku were generated than ultimately required. Since the questionnaire required three haiku per season for each model, generation continued until at least three outputs satisfying both formal constraints were obtained for each season. Constraint satisfaction required (i) canonical 5-7-5 mora structure and (ii) inclusion of an appropriate seasonal word (kigo). Outputs failing either criterion were excluded from the evaluation pool. Across models, successful adherence rates varied substantially. Acceptance rates and corresponding loss rates are reported in Appendix~\ref{app:constraint_adherence}. These results indicate that few-shot prompting improved controllability, but structural reliability remained model-dependent.

To support reproducibility, the complete experimental pipeline, including preprocessing scripts, prompting templates, generation code, statistical analyses, and the human evaluation questionnaire, is publicly available in the accompanying GitHub repository: \href{https://github.com/Livia020799/haiku_thesis}{Haiku Thesis}.

\section{Evaluation Methodology}

Since the aim of this study is to examine how AI-generated haiku are perceived and attributed by human readers, the primary evaluation was based on human judgment. To this end, AI-generated haiku were combined with those written by humans and presented in a questionnaire designed to examine how readers perceived, evaluated, and attributed authorship to the poems. Alongside this human-centred evaluation, the study employed exploratory AI-based assessments to complement the questionnaire results with structural, distributional, and model-based perspectives.

\subsection{Human Evaluation Design}

As creativity ultimately depends on human perception and interpretation, this study's central evaluation relied on a questionnaire-based human assessment. Its design was informed by prior works in computational poetry evaluation \cite{yan2016ipoet, hirota2018seqgan, tomizawa2025haiku}.
A pilot study ($N = 4$–$5$) was conducted at the Language Processing Laboratory of Shibaura Institute of Technology to refine the questionnaire design. Based on participant feedback, rating definitions were clarified, impression categories were adjusted to improve interpretability, and the original 24-haiku questionnaire was divided into two shorter 12-haiku versions. Following these adjustments, the finalized questionnaire was distributed to 144 participants across multiple Japanese universities, including Atomi University and Shibaura Institute of Technology. Most participants were native Japanese speakers enrolled in humanities (e.g., literature, psychology) and STEM programs. In addition, a small number of exchange students participated, with varying levels of Japanese proficiency. Data collection was concluded on December 3rd, 2025. At the beginning of the questionnaire, before the haiku evaluation tasks, participants provided background information including Japanese language proficiency, academic department or field of study, and prior exposure to haiku (e.g., reading frequency or composition experience). These variables were later used to examine whether participant characteristics influenced recognition performance.

Participants were asked to rate each haiku on seven dimensions (impression ratings), using a Likert scale going from 1 up to 5: \textbf{fluency} (grammatical and syntactic), \textbf{haiku-like wording}, \textbf{poeticness} (imagery and emotion), \textbf{coherence across lines}, \textbf{understandability / meaningfulness}, \textbf{favourability} (overall liking), and \textbf{unexpectedness} (surprise/novelty). In addition to aesthetic ratings, participants completed an authorship-attribution task, indicating whether they believed that each haiku was written by a human or generated by AI. This task was not used as a standalone test of machine creativity, but as a measure of perceived authorship to be compared with impression ratings and correct source recognition. Each questionnaire consisted of a sequence of haiku evaluation tasks. For every poem, participants completed the judgments described above: impression ratings, authorship attribution, confidence rating, and theme relevance.

To prevent participants from inferring the distribution of sources or anticipating the origin of the poems, AI-generated and human-written haiku were randomly mixed within each questionnaire. Human-written haiku were selected from curated online prize/award winning haiku contests hosted by Sophiakai \footnote{\url{https://www.sophiakai.gr.jp/news/others/2025080102.html}}, Haiku Nippon \footnote{\url{https://haikunippon.net/prize/2023w.html}} \footnote{\url{https://haikunippon.net/prize/2022au.html}} \footnote{\url{https://haikunippon.net/prize/2022summer.html}} \footnote{\url{https://haikunippon.net/prize/2022sp.html}}, and the EJC Foundation \footnote{\url{https://ejca.org/Spring-Haiku-Contest-2022}} to ensure that the poems reflected authentic contemporary haiku style and comparable quality to the generated outputs. This selection strategy was intended to avoid comparing AI-generated haiku with uncurated or low-quality human examples, although it does not imply that the human and AI poems were matched for all possible stylistic or aesthetic dimensions. 

For each of the six evaluated language models, three haiku were generated for each season (spring, summer, autumn, and winter), yielding 12 AI-generated haiku per model and 72 AI-generated haiku overall. A reference set of 12 human-written haiku, comprising three poems per season, was compiled from prize-winning entries in contemporary haiku competitions.

To facilitate classroom-based data collection while limiting participant burden, the evaluation dataset was organized into 12 questionnaire sets. For each language model, one questionnaire combined the three spring and three summer AI-generated haiku with the corresponding six human-written spring and summer haiku, whereas a second questionnaire combined the three autumn and three winter AI-generated haiku with the corresponding six human-written autumn and winter haiku.

Consequently, each questionnaire contained 12 randomly ordered haiku: six AI-generated and six human-written. Each participant completed only one questionnaire set. The final evaluation involved 144 participants, with each questionnaire set evaluated independently by a group of 12 participants. Participants were informed that both AI-generated and human-written haiku were present, but were not told the source of individual poems.

\begin{table}[H]
\centering
\caption{Representative GPT-5-generated haiku included in the evaluation dataset.}
\label{tab:gpt5_examples}
\small

\begin{tabularx}{\textwidth}{l >{\centering\arraybackslash}p{3.4cm} X}
\toprule
\textbf{Season} & \textbf{Japanese} & \textbf{English translation} \\
\midrule

Spring &
\begin{tabular}[c]{@{}c@{}}
春の風 \\
川面に揺れて \\
柳かな
\end{tabular}
&
\begin{tabular}[c]{@{}l@{}}
The spring breeze, \\
Swaying the river surface, \\
Weeping willow.
\end{tabular}
\\[1ex]

Summer &
\begin{tabular}[c]{@{}c@{}}
夕立に \\
子らの笑声 \\
消えにけり
\end{tabular}
&
\begin{tabular}[c]{@{}l@{}}
In the evening summer shower, \\
the children's laughter \\
has vanished.
\end{tabular}
\\[1ex]

Autumn &
\begin{tabular}[c]{@{}c@{}}
月見れば \\
古き井戸より \\
影のこゑ
\end{tabular}
&
\begin{tabular}[c]{@{}l@{}}
When you look at the moon, \\
From the old well, \\
A shadow's voice.
\end{tabular}
\\[1ex]

Winter &
\begin{tabular}[c]{@{}c@{}}
霜柱 \\
草履の跡に \\
朝ひかり
\end{tabular}
&
\begin{tabular}[c]{@{}l@{}}
Frost columns, \\
in the footprints of zori, \\
morning light.
\end{tabular}
\\

\bottomrule
\end{tabularx}
\end{table}

The pilot study also suggested that participants primarily relied on imagery richness, emotional depth, rhythmic flow, and phrasing naturalness when making authorship judgments.

\subsection{AI-based Evaluation}

In addition to human evaluation, we conducted two exploratory LLM-as-judge analyses to examine how a language model acting as an evaluator would classify and evaluate the haiku.

The first analysis was conducted with GPT-5 \cite{openai2025gpt5} during the pilot phase. GPT-5 was used to obtain an initial qualitative indication of how an advanced LLM would rate a subset of haiku across the same impression-rating dimensions used in the questionnaire (fluency, haiku-like wording, poeticness, coherence, understandability, favourability, and unexpectedness), and to provide brief justifications for its ratings.
The second analysis employed Grok \cite{xai2025grok4} for the authorship-classification task. Grok was selected as an external evaluator because GPT-5 and Gemini 2.5 were already included among the generation models, whereas Grok provided an additional strong contemporary model not used to generate the evaluated haiku. It was asked to classify each haiku in the evaluation set as either human-written or AI-generated and to justify its decision. Confusion matrices were constructed to compare Grok's classifications with ground-truth labels and with human recognition performance. The textual justifications were analyzed using an exploratory qualitative content-analysis procedure. Responses were manually reviewed, recurrent attribution cues were coded inductively and grouped into broader thematic categories, such as seasonality, mechanical phrasing, coherence, interpretability, and poetic depth. A detailed coding protocol with example outputs is provided in Appendix~\ref{app:grok-qualitative}, while the full evaluation setup and quantitative results are reported in Appendix~\ref{app:ai_evaluation_details}.

\subsection{Statistical Analysis}

To analyze recognition performance and impression-based attribution patterns, we employed a combination of regularized logistic regression and Bayesian generalized linear mixed models (GLMMs). Logistic regression models with L2 regularization were used for predictive analyses involving participant-level background variables and impression-rating predictors. Model robustness was evaluated using stratified cross-validation and grouped cross-validation procedures, depending on the structure of the analysis. Recognition performance was evaluated using accuracy and cross-validated Area Under the ROC Curve (AUC), while odds ratios were reported as descriptive effect-size indicators.

To account for repeated-measures dependencies in the questionnaire data, Bayesian binomial mixed-effects models were fitted using \texttt{statsmodels}' \texttt{BinomialBayesMixedGLM}. The models included random intercepts for both respondents and haiku items, reflecting the crossed structure of the dataset in which each participant evaluated multiple haiku and each haiku was evaluated by multiple participants. Bayesian estimation was performed using the default prior settings implemented in \texttt{statsmodels}, and model fitting was conducted using variational Bayes estimation. Given the exploratory nature of the study and the relatively limited participant sample, inferential results should be interpreted cautiously. In particular, near-chance AUC values are interpreted not simply as failed classification, but as theoretically informative evidence that aesthetic plausibility may exceed stable human discriminability in constrained poetic settings.

\section{Results}

\subsection{Recognition Accuracy}

\begin{table}[H]
\centering
\caption{Overall recognition accuracy by generation model.}
\label{tab:recognition_accuracy}

\begin{tabular}{lcc}
\toprule
\textbf{Model} & \textbf{Recognition accuracy} & \textbf{Interpretation} \\
\midrule
GPT-5       & 0.493 & Chance-level \\
Gemini 2.5  & 0.503 & Chance-level \\
StableLM-7B & 0.524 & Chance-level \\
LLM-JP      & 0.590 & Moderate detectability \\
Gemma-2B    & 0.611 & Moderate detectability \\
LLaMA-2     & 0.670 & Highest detectability \\
\bottomrule
\end{tabular}

\end{table}

Participants' ability to distinguish between AI-generated and human-written haiku was evaluated through recognition accuracy in the questionnaire-based authorship judgment task. Table~\ref{tab:recognition_accuracy} summarizes overall recognition performance across the six generation models. Performance varied substantially across models, with several systems remaining close to chance. GPT-5 and LLaMA-2 provide two representative cases of this contrast. For GPT-5, classification was nearly balanced and close to chance: 75 human-written haiku were correctly identified as human, 69 human-written haiku were misclassified as AI-generated, 77 AI-generated haiku were misclassified as human-written, and 67 AI-generated haiku were correctly identified as AI-generated. This corresponds to an overall recognition accuracy of 0.493. By contrast, LLaMA-2 showed clearer detectability: 100 human-written haiku were correctly identified as human, 44 were misclassified as AI-generated, 51 AI-generated haiku were misclassified as human-written, and 93 were correctly identified as AI-generated, corresponding to an overall recognition accuracy of 0.670.

\begin{comment}
Figures~\ref{fig:confusion_gpt5} and~\ref{fig:confusion_llama} show confusion matrices for GPT-5 and LLaMA-2, two representative models illustrating this contrast:

\begin{figure}[H]
\centering
\includegraphics[width=0.6\columnwidth]{gpt5_confusion.png}
\caption{Confusion matrix for GPT-5}
\label{fig:confusion_gpt5}
\end{figure}

\begin{figure}[H]
\centering
\includegraphics[width=0.6\columnwidth]{llama_confusion.png}
\caption{Confusion matrix for LLaMA-2}
\label{fig:confusion_llama}
\end{figure}
\end{comment}

The closed-source models, GPT-5 and Gemini 2.5, as well as the open-source model StableLM-7B, produced recognition accuracies that were consistent with chance-level classification. Similar near-chance patterns were observed for Gemini 2.5 (recognition accuracy = 0.503) and StableLM-7B (recognition accuracy = 0.524), where classification patterns remained nearly symmetric. By contrast, the remaining three open-source models demonstrated moderate detectability. Gemma-2B achieved an overall accuracy of 0.611, while LLM-JP reached 0.590. LLaMA-2 demonstrated the strongest performance (recognition accuracy = 0.670), with substantially higher correct classification rates observed for both AI and human classes.

\begin{figure}[H]
\centering
\includegraphics[width=0.75\textwidth]{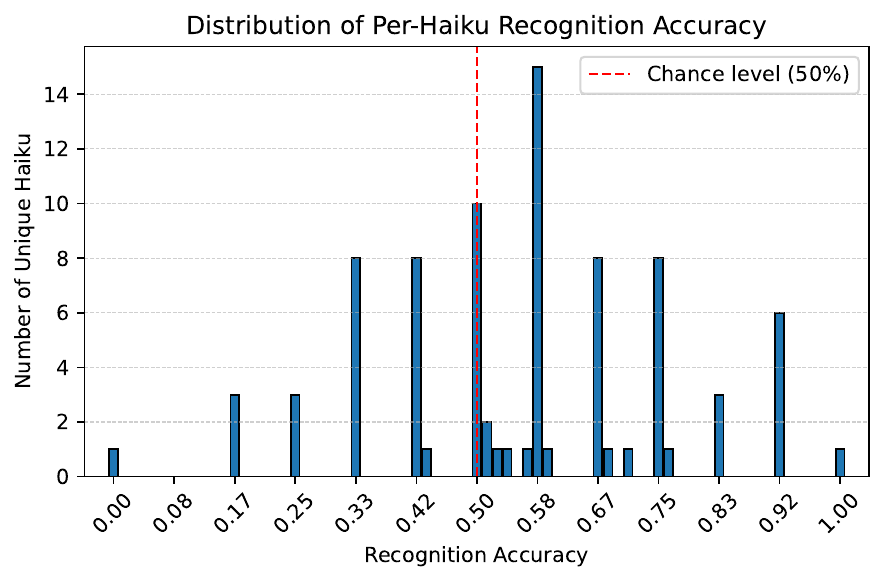}
\caption{Distribution of recognition accuracy across the 84 unique haiku included in the evaluation set.}
\label{fig:item_accuracy}
\end{figure}

Despite these differences, item-level analyses revealed substantial variability across individual haiku. Figure~\ref{fig:item_accuracy} shows the distribution of recognition accuracy across the 84 unique haiku included in the evaluation set, comprising 72 AI-generated haiku and 12 human-written haiku. For human-written haiku reused across model-specific questionnaires, recognition accuracy was averaged across all corresponding participant responses.

For each haiku, recognition accuracy represents the proportion of correct human authorship judgments aggregated across participants. The dashed vertical line indicates chance-level performance (50\% recognition accuracy). Although many haiku exhibited recognition accuracies near chance level, the distribution was broad, indicating considerable variability across items.

Recognition performance for individual haiku often remained close to chance, suggesting that participants generally found it difficult to reliably distinguish AI-generated from human-written poems on an item-by-item basis. A small subset of haiku achieved relatively high recognition accuracy, whereas another subset exhibited below-chance recognition accuracy, indicating that participants consistently misidentified their authorship.

Overall, recognition accuracy remained close to chance for half of the evaluated models, while item-level analyses suggest that detectability depends primarily on item-specific stylistic or linguistic cues rather than stable model-level characteristics.

\subsection{Impression Ratings}

This section reports the seven impression ratings descriptively, that is, as overall average evaluations across poems and models. These descriptive summaries should be distinguished from the inferential analyses reported in Section 5.4, where the same rating dimensions are used as predictors of correct recognition and perceived authorship. Across models and seasonal splits, impression ratings cluster in the mid-to-high range of the Likert scale (approximately 3-4). This indicates that both AI-generated and human-written haiku were generally perceived positively on most evaluative dimensions. A visualization of the overall mean ratings across the seven dimensions is provided in Appendix~\ref{app:overall_impressions}.

%\begin{figure}[H]
%\centering
%\includegraphics[width=0.9\columnwidth]{overall_impression_ratings.pdf}
%\caption{Overall impression ratings across evaluation dimensions}
%\label{fig:impression_ratings}
%\end{figure}

The highest ratings were observed for \textit{Fluency}, \textit{Haiku-like wording}, and \textit{Coherence}, all of which consistently fall in the upper part of the observed range. This suggests that many poems, regardless of authorship, were perceived as grammatically natural, stylistically plausible, and internally coherent. \textit{Poeticness}, \textit{Understandability}, and \textit{Favourability} also received moderately positive scores, indicating generally favourable aesthetic evaluations. By contrast, \textit{Unexpectedness} received the lowest average rating, remaining closer to the midpoint of the scale. This suggests that participants perceived many haiku as competent and acceptable, but less often as surprising or strongly novel. AI-generated and human-written haiku received largely similar impression ratings across these dimensions. Confidence ratings, however, showed a different pattern. For several models, particularly StableLM-7B and LLM-JP, confidence values shifted towards the lower end of the scale despite relatively high impression ratings. In other words, participants often considered the poems well-formed or aesthetically acceptable while remaining uncertain about whether they had been written by a human or generated by AI.

Model-specific differences are also evident. Outputs from GPT-5, Gemini 2.5 and Gemma-2B demonstrate relatively consistent mid-range ratings with limited polarization among respondents. By contrast, StableLM-7B shows a clearer divergence between impression ratings and confidence scores. Overall, impression ratings remain concentrated in the mid-to-high range across most dimensions, whereas confidence values tend to be lower and more variable across models. The impression-rating analyses provide further insight into this phenomenon. Ratings were generally high for fluency, haiku-like wording and coherence. This suggests that both AI-generated and human-written haiku were generally considered linguistically acceptable and stylistically plausible. In contrast, confidence ratings were often lower, indicating that, while participants frequently judged the poems as well-formed, they remained uncertain about their origin, suggesting that linguistic quality alone was not sufficient for reliable authorship attribution.

\subsection{Participant Background and Recognition Ability}

A logistic regression analysis was conducted to examine whether participant characteristics influence recognition accuracy, using respondent-level predictors: Japanese language proficiency, academic background, and prior exposure to haiku (haiku experience and composition). To improve interpretability and reduce sparsity in the open-ended department responses, the departments were aggregated into broader categories: STEM, Social Sciences/Humanities, Psychology, and Unknown.

Among the participant-level variables examined, Japanese language proficiency showed the clearest association with recognition performance. Compared to native speakers, several levels of non-native proficiency correspond to reduced odds of above-chance recognition. In contrast, variables related to haiku exposure, such as reading frequency or prior composition experience, demonstrate weak and inconsistent effects. Unexpectedly, a positive association was observed for participants from STEM disciplines compared to those from the humanities and social sciences. However, given the exploratory nature of the analysis and potential confounding factors (such as familiarity with AI systems or differences in response strategies), this effect should be interpreted cautiously. One possible explanation is that participants from STEM disciplines may have had greater prior exposure to AI systems, text-generation tools, or discussions surrounding generative models, making them more attentive to recurrent stylistic cues often associated with machine-generated text. Another possibility is that these participants approached the task as an analytic classification problem, relying more heavily on pattern detection than on purely aesthetic intuition. However, the present study did not directly measure familiarity with AI tools or decision strategies. Therefore, this association should be interpreted as exploratory rather than causal.

To assess whether the observed patterns were primarily driven by language proficiency, two additional models were estimated: a language-only model including Japanese proficiency as the sole predictor, and a no-language model including all background variables except language proficiency. Predictive performance was evaluated using a three-fold cross-validated Area Under the Curve (AUC). The full regression model achieved an AUC of 0.416, the language-only model reached 0.456, and the no-language model reached 0.432. These values all remain close to chance level, indicating that participant background variables provide limited discriminative power for predicting the recognition of AI-generated versus human-written haiku above chance levels. Across model specifications, coefficient patterns are strongest for Japanese proficiency contrasts relative to mother-tongue speakers, while effects for haiku exposure and department category are smaller and less consistent. Consistent with this, cross-validated AUC values are near chance in all models. These weak predictive results suggest that participant background characteristics alone provide limited explanatory power for authorship recognition, reinforcing the interpretation that recognition performance is highly item-dependent rather than strongly determined by stable participant-level traits.

\subsection{Perceived Authorship vs.\ Correct Recognition}

Whereas Section 5.2 summarizes the seven impression dimensions descriptively, the present section uses these same dimensions as explanatory variables. In addition to the seven impression ratings, theme relevance was included as a separate predictor because it was collected independently in the questionnaire.

\begin{figure}[H]
\centering
\includegraphics[width=0.7\columnwidth]{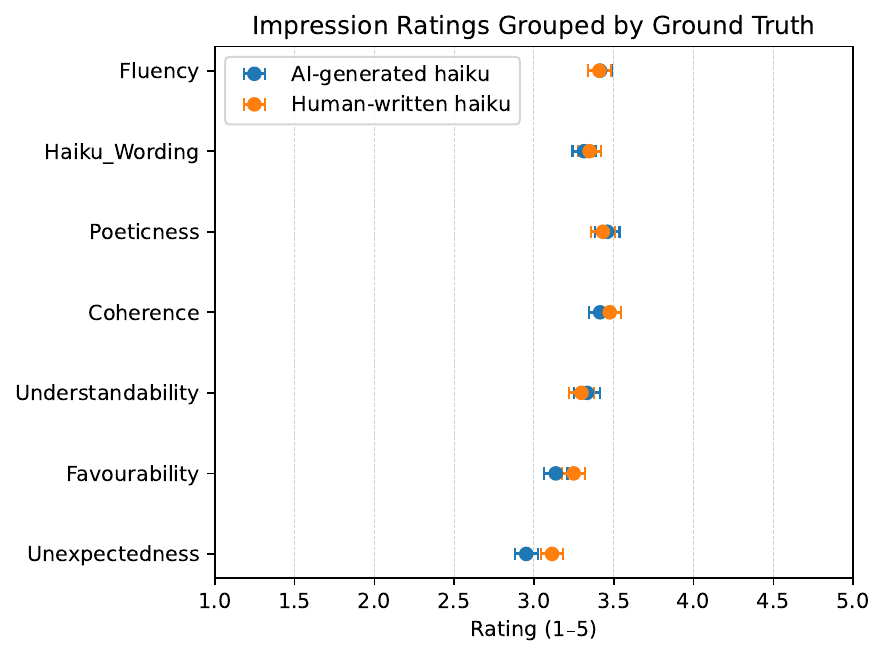}
\caption{Mean impression ratings (±95\% confidence intervals) grouped according to the true authorship of each haiku}
\label{fig:ground_truth}
\end{figure}

\begin{figure}[H]
\centering
\includegraphics[width=0.7\columnwidth]{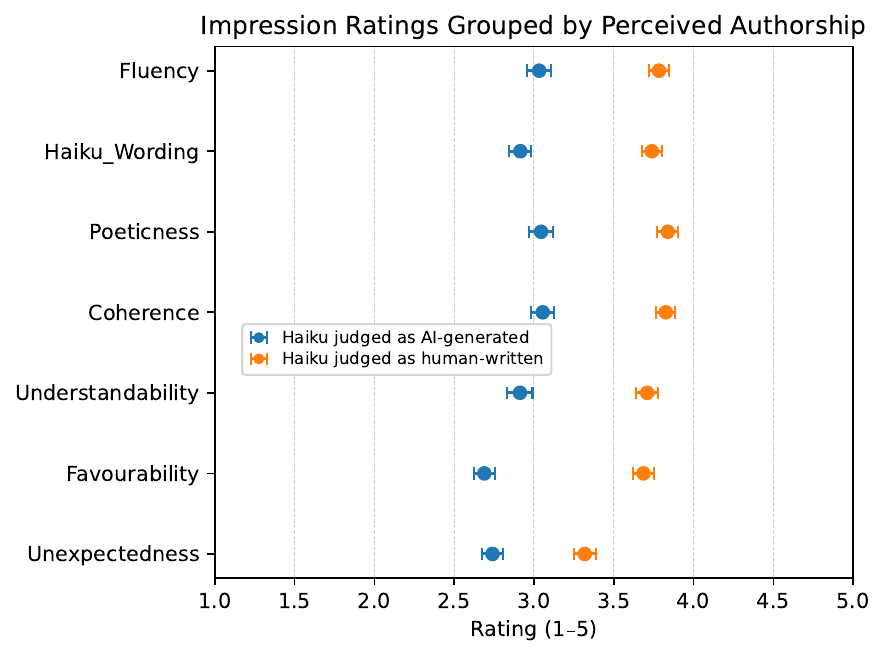}
\caption{Mean impression ratings (±95\% confidence intervals) grouped according to participant-perceived authorship}
\label{fig:perceived_authorship}
\end{figure}

To investigate the relationship between impression ratings and authorship judgments, Bayesian generalized linear mixed models (GLMMs) were used. The first model examined whether impression cues could predict \textit{correct authorship recognition} (i.e., whether participants correctly identified the true source of the haiku), while two additional models analyzed \textit{perceived authorship} (judged as human or AI). The results revealed a clear distinction between recognition and perception. Figures~\ref{fig:perceived_authorship} and~\ref{fig:ground_truth} illustrate how impression ratings differ when grouped by the true source of the haiku and when grouped by perceived authorship.

When it comes to predicting \textit{correct recognition}, impression ratings only show weak association with recognition accuracy. Higher favourability ratings are associated with slightly lower odds of correct classification (OR = 0.83), suggesting that haiku perceived as more pleasant are more likely to be misclassified. Conversely, understandability (OR = 1.08), unexpectedness (OR = 1.08), theme relevance (OR = 1.78) and fluency (OR = 1.05) demonstrate modest positive correlations with accurate recognition. Poeticness shows a weak negative association (OR = 0.96), while coherence (OR = 0.98) and haiku wording (OR = 0.99) have little effect. Overall predictive performance remains near chance, with a cross-validated AUC of 0.505.
Although several predictors show associations with recognition performance, the observed effect sizes remain relatively small overall. This suggests that no single impression dimension strongly determines correct authorship recognition in isolation.

By contrast, the same cues provide a strong indication of perceived authorship. The model that predicts whether a haiku is judged to have been written by a human achieves a cross-validated area under the curve (AUC) of 0.793. In contrast to the weak associations observed for correct recognition, the stronger effect sizes observed for perceived authorship indicate that impression ratings are more predictive of attribution heuristics than of true source identification.
The strongest predictors are favourability (OR = 1.72), haiku-like wording (OR = 1.36) and theme relevance (OR = 1.33). Unexpectedness also has a positive effect, meaning that poems that are more surprising are more likely to be perceived as having been written by a human. Smaller positive effects were observed for fluency, coherence and understandability, while poeticness showed a slight negative association.

The complementary model predicting AI judgments exhibits the same pattern in reverse: lower favourability, poorer wording quality and weaker thematic fit are all associated with a higher likelihood of AI attribution. Importantly, impression ratings were more closely linked to \textit{perceived authorship} than to \textit{true authorship}. Participants consistently gave higher ratings to haiku that they thought were written by humans, and lower ratings to those that they thought were generated by AI. In other words, aesthetic impressions appeared to influence attribution judgements, even when they did not reflect the poem's actual authorship.

\begin{comment}

\subsection{Exploratory AI-based Evaluation}

To complement the human evaluation, two exploratory LLM-as-judge analyses were conducted.
GPT-5 rated a subset of haiku on the same impression dimensions used in the questionnaire and generally assigned mid-to-high scores, often referring to imagery and seasonal coherence.
Grok \cite{xai2025grok4} was instead asked to classify each poem as human-written or AI-generated. Consistent with the human results, it struggled to distinguish outputs from stronger models (GPT-5, Gemini, Gamma) but showed clearer separability for smaller models such as Gemma-2B, Instruct4 and LLaMA.

\end{comment}

\section{Discussion}

The results suggest that participants generally struggled to distinguish reliably between AI-generated and human-written haiku, and that impression ratings were more closely linked to perceived authorship than to the poems' true source. This pattern aligns with prior research showing that contextual information and attribution framing can influence aesthetic evaluation of creative artifacts. Studies of AI-generated visual art have demonstrated that perceived authorship can shape both moral and aesthetic judgments independently of the artifact itself \cite{BARA2025106063,CHIARELLA2022107406}. The present results extend this observation to computational poetry, suggesting that evaluative impressions may function primarily as heuristics for attribution rather than as reliable diagnostic signals.

From the perspective of computational creativity theory, these findings emphasise the distinction between creative evaluation and creative attribution. As noted by Colton and Wiggins, judgements of computational creativity often depend on how observers interpret the source and process behind an artefact \cite{colton2012computational}. This experiment shows that, even when attempting to infer authorship under blind conditions, observers' decisions are influenced by aesthetic expectations that do not necessarily reflect the text's true origin. In this sense, authorship attribution seems to reflect perceptual strategies rather than diagnostic cues. Participants rely on features such as favourability, the naturalness of the wording, and thematic appropriateness when reasoning about authorship. However, these features do not reliably distinguish between AI-generated and human-written haiku. Rather, they primarily influence whether a poem feels human- or AI-generated.

Given that recognition accuracy for GPT-5, Gemini 2.5, and StableLM-7B remained close to chance level, the results do not support a simple interpretation in which participants systematically attributed lower-quality poems to AI. If perceived poor quality alone had driven AI attribution, recognition accuracy for AI-generated haiku would likely have been consistently above chance across models. Instead, performance was heterogeneous: while GPT-5, Gemini 2.5, and StableLM-7B remained near chance, other systems such as Gemma-2B, LLM-JP, and especially LLaMA-2 showed moderate detectability. This pattern suggests that participants could not reliably distinguish AI-generated from human-written haiku on the basis of quality cues alone. Rather, perceived authorship and aesthetic evaluation appear to interact, with judgments reflecting broader expectations about what AI-generated poetry should look like, alongside model-specific textual cues. The tentative association between STEM background and recognition performance also raises an important methodological question: the ability to detect AI-generated creative text may depend not only on literary sensitivity, but also on technological familiarity and task framing. If so, authorship recognition tasks may partially measure participants’ mental models of AI systems rather than sensitivity to poetic quality itself.

One possible explanation for this pattern relates to how people conceptualize AI-generated text. The cues that participants relied on may reflect stereotypical expectations about machine-generated language, rather than the actual properties of contemporary AI systems. Participants tended to associate AI authorship with characteristics such as poorer word choice, lower aesthetic appeal, and reduced thematic coherence. However, current large language models are capable of producing highly fluent, stylistically appropriate text, including structured poetic forms such as haiku. This suggests that participants may rely on outdated expectations about AI-generated language, leading them to attribute AI authorship based on perceived stylistic deviations rather than the poem’s true source. In this sense, authorship judgments may be guided less by objective textual cues and more by heuristic assumptions about how machines are expected to write, which can lead to systematic misclassification when modern AI systems produce outputs that match or exceed human stylistic performance.

\section{Conclusions}

This study investigated whether readers could distinguish between AI-generated and human-written Japanese haiku and how aesthetic impressions relate to authorship attribution. Recognition accuracy remained close to chance for several contemporary language models, although some systems showed moderate detectability. At the same time, recognition varied substantially across individual haiku, indicating that detectability is strongly item-dependent.

The central finding is a dissociation between aesthetic evaluation and correct authorship recognition. Impression ratings were substantially more predictive of whether a poem was \emph{perceived} as human-written than of whether its true authorship was identified correctly. Aesthetic cues such as favourability, haiku-like wording, and thematic relevance therefore appear to function primarily as attribution heuristics rather than as reliable diagnostic signals of true authorship.

These findings have broader implications for the evaluation of generative AI systems. As language models become increasingly capable of producing fluent and stylistically plausible creative text, perceived quality and true authorship may become progressively less aligned. Human evaluation remains valuable for understanding perceived aesthetic quality, but aesthetic judgments alone may become less reliable as a mechanism for identifying the source of creative artifacts.

Future work should investigate more diverse human-written corpora, additional language models and decoding strategies, contextual framing effects, and collaborative human--AI creative settings. Such extensions could clarify how perceptions of creativity, quality, and authorship evolve as generative systems become increasingly integrated into creative practice.

\section{Limitations}

Several limitations should be considered when interpreting these results. First, the study examines a specific constrained setting: Japanese seasonal haiku generated under a few-shot prompting framework. The findings therefore should not be generalized directly to unconstrained creative writing or to other artistic domains.
Second, the human evaluation involved 144 university participants from a limited set of Japanese institutions. Although most participants were native Japanese speakers and represented both humanities and STEM disciplines, the sample does not represent the broader Japanese population.
Third, the human-written comparison set consisted exclusively of prize-winning haiku from contemporary competitions. This provided a curated reference set and avoided comparison with uncurated or low-quality human examples, but it may also introduce selection bias because the human reference poems represent a particularly successful subset of contemporary haiku.
Fourth, the study compares six specific generative models under a common prompting and decoding framework. Recognition patterns may change with different models, model versions, prompts, decoding strategies, or future systems.
Fifth, the statistical analyses are exploratory. Although mixed-effects models were used to account for repeated observations across respondents and haiku items, the relatively limited participant sample and near-chance predictive performance in several analyses require cautious interpretation.
Finally, the LLM-as-judge analyses reported in the appendices are exploratory rather than confirmatory. In particular, Grok's textual explanations should be interpreted as model-generated rationales rather than as transparent evidence of the model's internal decision process.

\clearpage

\begin{appendices}

\section{Constraint Adherence and Loss Rates}
\label{app:constraint_adherence}

Table~\ref{tab:constraint_adherence} reports the proportion of generated candidate haiku satisfying both formal constraints: (i) canonical 5-7-5 mora structure and (ii) inclusion of an appropriate seasonal word (kigo). The loss rate corresponds to the proportion of rejected candidates.

\begin{table}[H]
\centering
\caption{Constraint adherence, acceptance rate, and loss rate across models.}
\label{tab:constraint_adherence}
\footnotesize
\setlength{\tabcolsep}{4pt}
\begin{tabular}{lcc}
\toprule
Model & Acceptance rate & Loss rate \\
\midrule
StableLM-7B & 62.5\% & 37.5\% \\
Gemma-2B & 83.3\% & 16.7\% \\
LLaMA-2 & 78.6\% & 21.4\% \\
LLM-JP & 100.0\% & 0.0\% \\
GPT-5 & 91.7\% & 8.3\% \\
Gemini 2.5 & 100.0\% & 0.0\% \\
\bottomrule
\end{tabular}
\end{table}

\section{Grok Qualitative Coding Protocol}
\label{app:grok-qualitative}

Grok was asked to classify each haiku as either human-written or AI-generated and to provide a short justification for its decision. The justifications were reviewed using an exploratory qualitative content-analysis procedure. In the first pass, recurrent attribution cues were identified inductively from the explanations. In the second pass, similar cues were grouped into broader thematic categories.

The final categories included: seasonality and kigo use; mechanical or formulaic phrasing; coherence and interpretability; poetic depth versus surface plausibility; imagery and emotional resonance; and naturalness of wording. The analysis was descriptive rather than confirmatory and was used to identify common reasoning patterns in Grok's classifications, not to establish a formal coding taxonomy.

This analysis should be interpreted as exploratory. No formal intercoder reliability was computed, and Grok's explanations are treated as model-generated rationales rather than transparent access to the model's internal decision process.

\section{Exploratory AI-based Evaluation}
\label{app:ai_evaluation_details}

\subsection{Overview}

To complement the human evaluation, two exploratory \emph{LLM-as-judge} approaches were considered:
(i) a direct scoring evaluation using ChatGPT-5, and 
(ii) an authorship classification task using Grok.
These analyses were not designed as formal benchmarks, but as exploratory probes to observe how automated systems behave when evaluating or classifying haiku.

\subsection{ChatGPT-5 as a Judge}

This analysis was conducted during the early pilot phase of the study as an exploratory qualitative probe, rather than as part of the main evaluation framework.
Using \texttt{ChatGPT-5} as a direct evaluator, a subset of five haiku was provided, and the model was asked to rate each poem across seven dimensions:

\begin{itemize}
\item Fluency (grammatical correctness)
\item Haiku-like wording
\item Poeticness (imagery and emotional impact)
\item Coherence across lines
\item Understandability / meaningfulness
\item Favourability (overall appreciation)
\item Unexpectedness (novelty or surprise)
\end{itemize}

Each dimension was evaluated on a 1-5 Likert scale. The model also provided short textual justifications and an overall score computed as an aggregate of the seven dimensions. The observed scores ranged between 2 and 5 across dimensions. For example, some haiku received consistently high scores (e.g., 4–5 across most dimensions), while others received lower evaluations (e.g., scores around 2–3 in coherence and understandability). The model provided short explanations for each score, typically referencing aspects such as imagery, coherence between lines, and clarity of meaning.
Given the small sample size (five haiku) and the exploratory nature of this test, these results are not used for comparison with human ratings, but are reported to document the behavior of the model in a structured evaluation setting.

\subsection{Grok as a Judge}

A second evaluation was conducted using \texttt{Grok}, which was prompted to classify each haiku as either \emph{human-written} or \emph{AI-generated}. The dataset consisted of the same evaluation set used in the human questionnaire, including balanced subsets for each model (12 human-written haiku and 12 AI-generated haiku per model). This ensures direct comparability between Grok’s classifications and human judgments on identical items. AI-generated haiku were produced by six models: GPT-5, Gemini 2.5, StableLM-7B, Gemma-2B, LLM-JP, and LLaMA-2.

Confusion matrices were computed for each model. For GPT-5, Gemini 2.5, and StableLM-7B, Grok classified all haiku as \emph{human-written}. This results in 12/12 correct classifications for human-written haiku, 0/12 correct classifications for AI-generated haiku, and an overall accuracy of 50\%, corresponding to chance-level performance due to the balanced dataset.

\begin{figure}[H]
\centering
\includegraphics[width=0.32\textwidth]{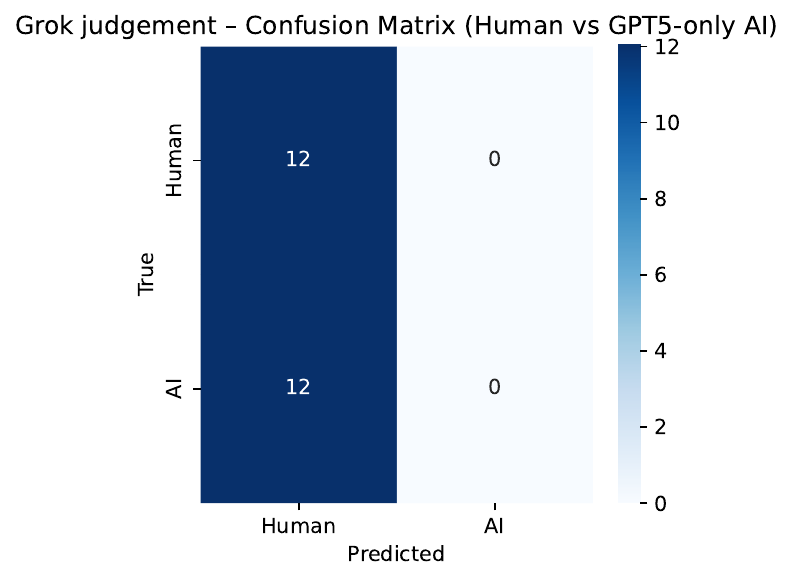}
\includegraphics[width=0.32\textwidth]{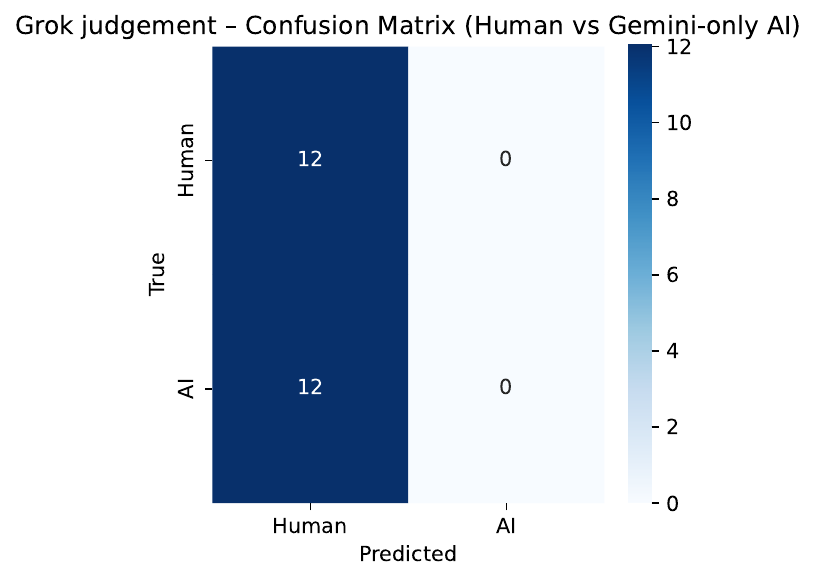}
\includegraphics[width=0.32\textwidth]{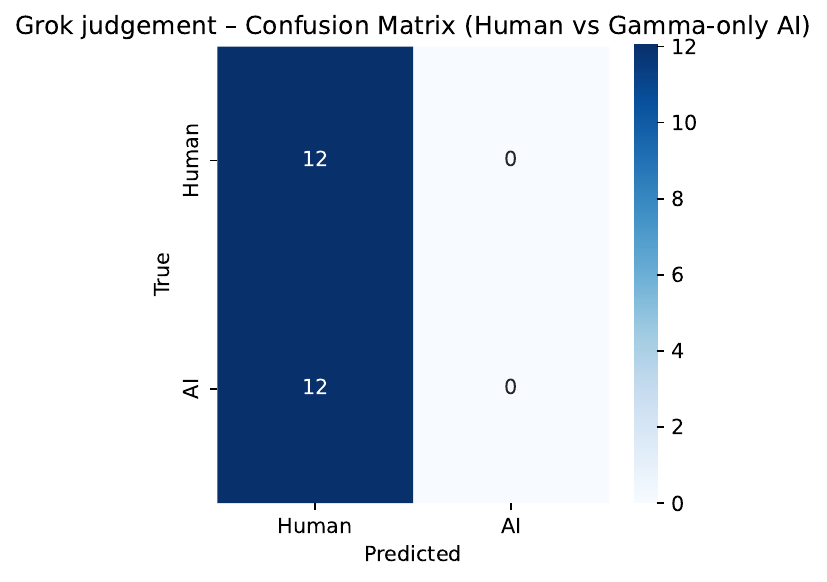}
\caption{Confusion matrices for GPT-5, Gemini 2.5, and StableLM-7B. All AI-generated haiku are classified as human-written, resulting in chance-level accuracy.}
\label{fig:grok_high_models}
\end{figure}

For the remaining models, Grok produced a mix of predictions. Gemma-2B achieved 20/24 correct classifications (83\% accuracy), LLM-JP achieved 17/24 correct classifications (71\% accuracy), and LLaMA-2 achieved 19/24 correct classifications (79\% accuracy).

\begin{figure}[H]
\centering
\includegraphics[width=0.32\textwidth]{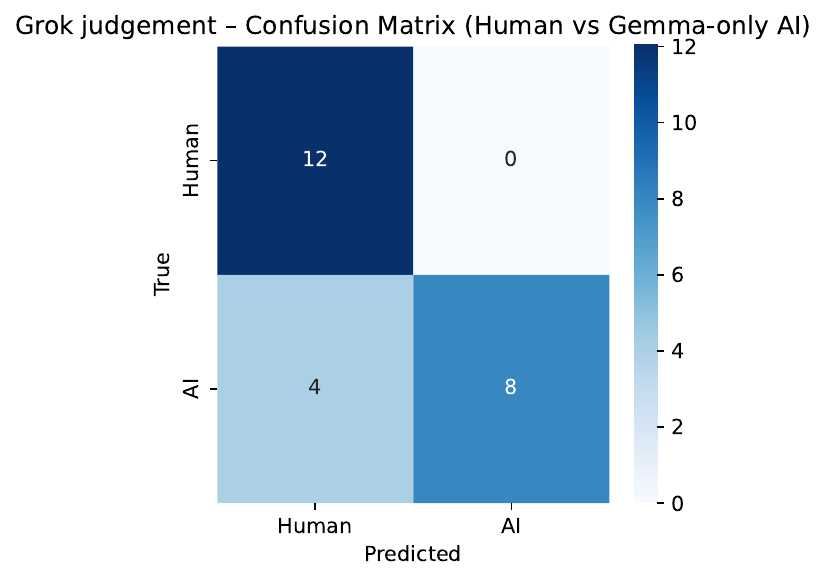}
\includegraphics[width=0.32\textwidth]{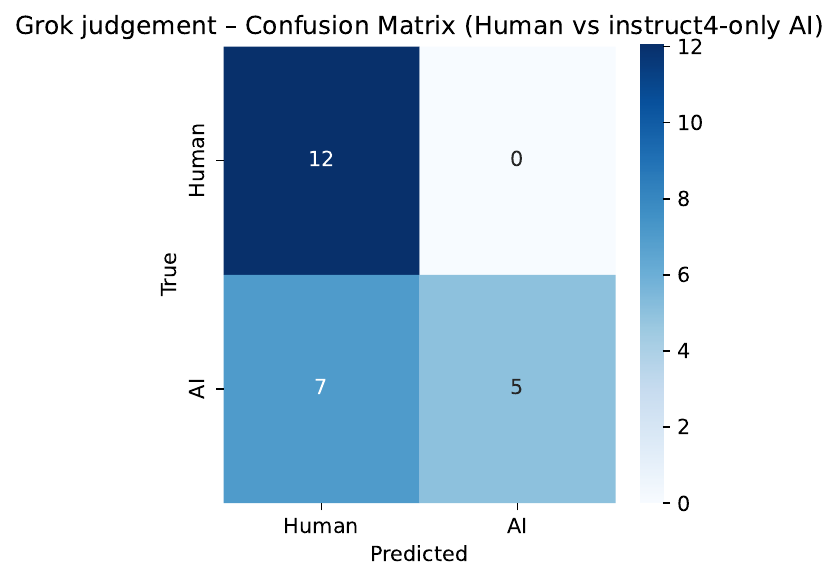}
\includegraphics[width=0.32\textwidth]{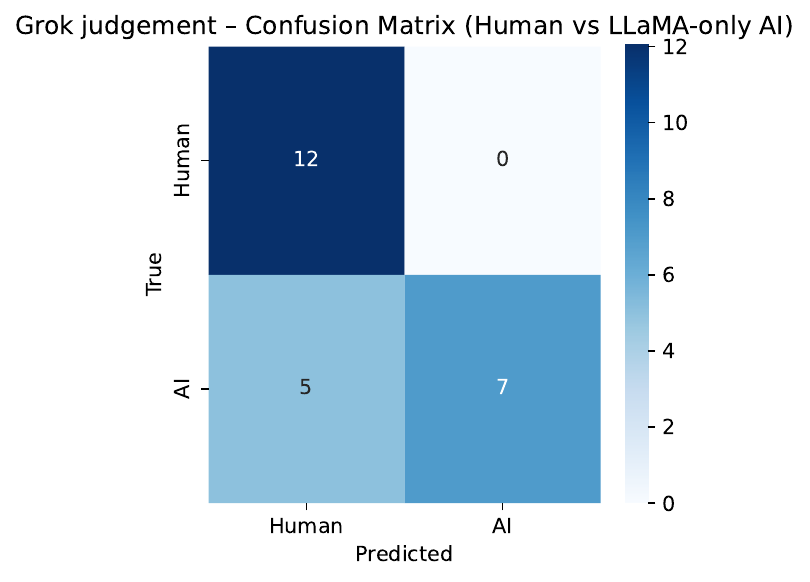}
\caption{Confusion matrices for Gemma-2B, LLM-JP, and LLaMA-2. In these cases, Grok correctly classifies a portion of AI-generated haiku, resulting in higher accuracy.}
\label{fig:grok_low_models}
\end{figure}

These values are directly derived from the confusion matrices shown in Figures~\ref{fig:grok_high_models} and~\ref{fig:grok_low_models}. In addition to the classification labels, Grok was asked to provide short justifications for each decision. A qualitative review of these explanations reveals recurring attribution patterns. Haiku classified as AI-generated are frequently described as exhibiting inconsistent or unclear seasonal references, repetitive or formulaic phrasing, weak or disjointed imagery, overly explicit or tautological descriptions, and forced or unclear metaphors. By contrast, haiku classified as human-written are typically associated with more coherent imagery, consistent use of seasonal elements, natural and balanced phrasing, and a more subtle or evocative emotional tone.

The results show two distinct behaviors. For GPT-5, Gemini 2.5, and StableLM-7B, Grok assigns all samples to the \emph{human} class, resulting in chance-level accuracy (50\%) due to the balanced dataset. For Gemma-2B, LLM-JP, and LLaMA-2, Grok correctly classifies a substantial portion of AI-generated haiku, leading to higher accuracy values.

Overall, this exploratory analysis highlights systematic differences in how Grok responds to outputs from different models. In particular, the results suggest a consistent pattern in which haiku generated by GPT-5, Gemini 2.5, and StableLM-7B are predominantly classified as human-written, whereas outputs from Gemma-2B, LLM-JP, and LLaMA-2 remain partially distinguishable. These observations are consistent with the variability in detectability reported in the main human evaluation, where recognition performance differed across models and was strongly dependent on individual items. The qualitative patterns observed in Grok’s justifications further suggest that classification decisions are often based on surface-level cues such as coherence, seasonal consistency, and phrasing naturalness, which may not always align with true authorship. However, given the limited sample size and exploratory design, these findings should be interpreted as indicative patterns rather than as generalizable estimates of model-level performance.

\section{Overall Impression Ratings}
\label{app:overall_impressions}

\begin{figure}[H]
\centering
\includegraphics[width=0.75\textwidth]{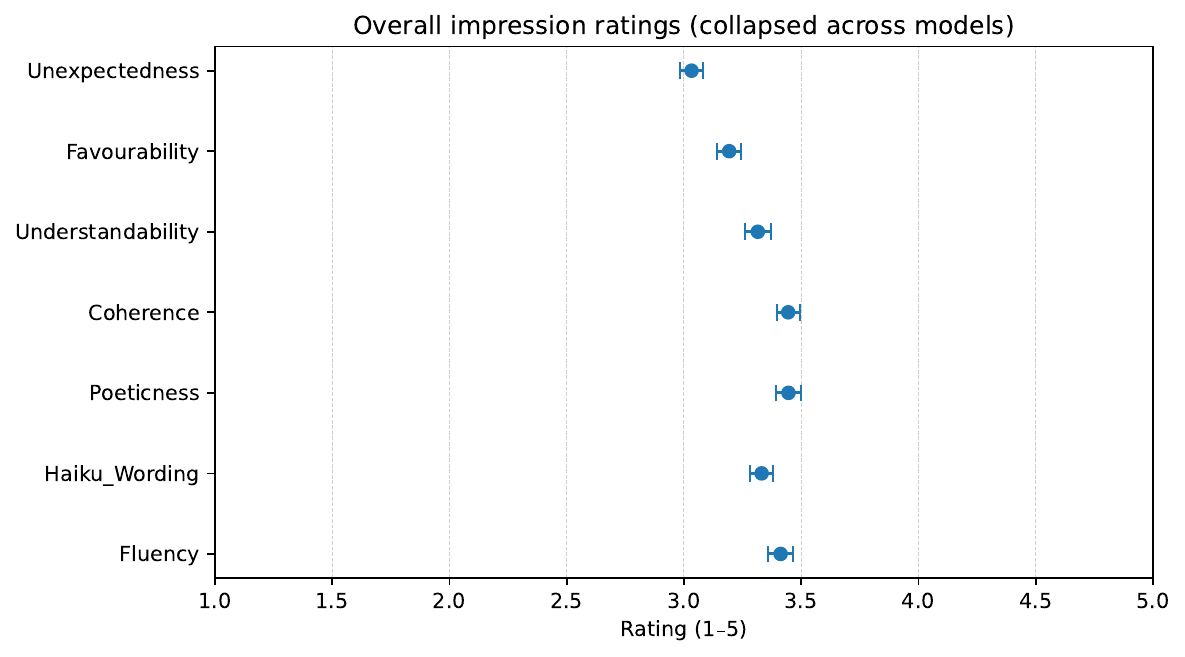}
\caption{Overall impression ratings across evaluation dimensions, collapsed across generation models.}
\label{fig:impression_ratings}
\end{figure}

Figure~\ref{fig:impression_ratings} provides the overall descriptive distribution of impression ratings across the seven evaluation dimensions, collapsed across generation models and authorship conditions.

\end{appendices}

\clearpage

\bibliographystyle{unsrt}
\bibliography{iccc}

\end{document}